\pdfoutput=1
\documentclass[11pt]{article}

\usepackage{acl}

\usepackage{times}
\usepackage{graphicx}
\usepackage{latexsym}

\usepackage[super]{nth}
\usepackage{enumitem}

\usepackage[T1]{fontenc}
\usepackage[utf8]{inputenc}

\usepackage{microtype}

\title{Assessing Digital Language Support on a Global Scale}

\author{Gary F. Simons \\
  SIL International \\
  \texttt{\small gary\_simons@sil.org} \\\And
  Abbey L. Thomas \\
  The University of Texas at Dallas \\
  \texttt{\small abbey.thomas@utdallas.edu} \\\And
  Chad K. White \\
  SIL International \\
  \texttt{\small chad\_white@sil.org} \\}

\begin{document}
\maketitle
\begin{abstract}
The users of endangered languages struggle to thrive in a digitally-mediated world. We have developed an automated method for assessing how well every language recognized by ISO 639 is faring in terms of digital language support. The assessment is based on scraping the names of supported languages from the websites of 143 digital tools selected to represent a full range of ways that digital technology can support languages. The method uses Mokken scale analysis to produce an explainable model for quantifying digital language support and monitoring it on a global scale. 

\end{abstract}

\section{Introduction}

%In the \nth{21} century context of accelerating globalization and technology on the one hand and the endangerment of non-dominant languages on the other, crossing the digital divide may be essential to the long-term survival of a language. We therefore want to be able to monitor and report what is happening in that regard.

The users of endangered languages struggle to thrive in a digitally-mediated world. The opportunities
afforded by digital technology differ drastically depending on the language being used. This has been
dubbed the ``digital language divide'' \citep{mikami_digital_2008,young_digital_2015,soria_what_2016,matsakis_bridging_2019}. As digital modes
of communicating and accessing information become increasingly necessary in daily life, lack of
digital language support (DLS) for a language means that its speakers must use other languages to participate in the global information society or be left out.

Linguists have been writing for decades about the role digital technology could play in language
revitalization  \citep{warschauer_technology_1998,buszard-welcher_can_2001,eisenlohr_language_2004,galla_indigenous_2009,holton_role_2011,cru_digital_2016}. Language technologists are recognizing the inequities facing the vast majority of the world's languages \citep{bird_decolonising_2020,blasi_systematic_2022} and are embracing the challenges of bringing greater equity in DLS \citep{joshi_unsung_2019,bapna_building_2022,edunov_teaching_2022}.

However, in a world where most people are multilingual and each language fits into its functional niche within an ecology of languages \citep{lewis_sustaining_2016}, full digital support for every language is not a realistic goal nor what those multilingual individuals are necessarily looking for \citep{bird-2022-local}. The goal of our research is to develop a method for measuring DLS in every language, so that it will be possible to provide an empirical view of the digital state of the world's languages and to observe the progress as so-called low-resource languages move toward crossing the digital language divide.

\section{Related Work}

Our primary inspiration has been the seminal work by \citet{kornai_digital_2013} on developing a method for
assessing the digital vitality of any language.  He proposes a four-way classification of languages as digitally Thriving, Vital, Heritage, or Still, ``roughly corresponding to the amount of digital communication that takes place in the language.'' His method 
harvests data from the Web, then uses supervised classification to automatically label all known languages. 
In practice, he adds a fifth level, Borderline, to represent languages that show signs of crossing the gap from Still to Vital. 
He and his colleagues have applied this method to the languages of India \citep{kornai_indian_2014},  the former Soviet Union \citep{kornai_new_2015}, and the Uralic family \citep{acs_digital_2017}.

In reviewing Kornai's method,  \citet{gibson_framework_2015,gibson_assessing_2016} focused on the huge gap between Still and Vital. He argues that two additional levels are needed to fill this gap: one for when  the needed elements (like a keyboarding solution) are in place for potential digital language use, and another for when digital language use is indeed taking off. 
% He proposes the names Latent and Emergent for these levels.
We follow Gibson's lead in adding two levels, but use names that achieve better congruence with the geometry of the S-curve model that emerges from our method (see Figure \ref{fig:sCurve}).

\section{Requirements}

Following Kornai's \citeyearpar{kornai_digital_2013} lead, we seek to develop an automated method for assessing digital language vitality that is based on feature data harvested from the Web. In this way, it can be run periodically to monitor changes in digital vitality for every language. We were motivated to develop an alternative to Kornai's method of analysis in order to meet three requirements:

\textit{Digital vitality should be orthogonal to non-digital vitality.}
We exclude features like population and language vitality from the feature data. Kornai notes that the EGIDS level as reported in Ethnologue \citep{lewis_assessing_2010,eberhard_ethnologue:_2022} is  “the best predictor of digital status.”  But digital vitality is distinct from non-digital vitality. For instance, our method reports the ``dead'' language Latin to be the \nth{80} most digitally vital language in the world. By contrast, Aimaq with nearly two million speakers is found to be digitally Still.

% In developing a working model of digital language vitality, we have been informed  by Hilbert’s \citeyearpar{hilbert_end_2011}  treatment of the digital divide as a study in the diffusion of innovations. He notes that studies of digital adoption typically look at two major components: the access to digital technology that is achieved and the usage that is actually made (Hilbert 2011:§3.4). In the case of a language, it achieves access to the digital medium when use of the language is supported in various digital technologies. That is, people can use the language in a digital medium because it is supported in things like fonts, keyboards, user interfaces, spelling checkers, and automatic translators. We refer to this as “digital language support.” The other major component of digital adoption is the kind and amount of digital use of the language. This may range from the preservation of digital records of the language in archives to the digital use of the language in everyday communication by members of the language community to the digital creation and dissemination of vast stores of information in the language. We refer to this as “digital language use.”

\textit{The assessments should be explainable.}
 A standard critique of  machine learning models based on black-box methods is that the models cannot explain why they produce the answers they do \citep{Arrieta2020ExplainableAI,miller_explanation_2019}.  \citet{kornai_digital_2013} bases his results on the majority outcome from 100 runs of a black-box model that yields a slightly different result each time. Users will be  more likely to trust results if they are deterministic and explainable.

\textit{The assessment scale should measure a single underlying trait.}
The data features used by \citet{kornai_digital_2013} covered a variety of digital uses. Some had to do with   quantifying the extent to which the language has been documented in digital archives by researchers. Others, like the sizes of Wikipedias, had to do with quantifying the extent of digital language use by the language community itself.  Still others looked at specific software products and recorded which languages they support. These strike us as three distinct traits, each of which should be assessed in its own right: digital language preservation, digital language use (DLU), and digital language support (DLS). 
Of these, the latter two are what speak to monitoring the digital vitality of a language as it moves toward crossing the digital language divide. DLU and DLS are distinct traits that should be assessed separately---speakers of unsupported languages may nevertheless use it digitally (for instance, making do in texting; see \citet{Eberhard_Mangulamas_2022}), while speakers of supported languages may choose to use digital resources in another language they know.

We have chosen to focus on DLS since the data for monitoring that phenomenon are openly accessible---the developers of digital tools are usually keen to advertise all of the languages they support. By contrast, data on  actual digital use is typically not shared on a language-by-language basis by the vendors concerned. A comparable effort to assess DLU on a global scale is much needed, though we anticipate that it will be significantly harder to acquire the needed data.

\section{Methodology}

The method we have adopted for building an explainable model of DLS is Mokken scale analysis \citep{mokken_theory_1971,schuur_mokken_2003}.  Mokken's method is a generalization of the more widely known Guttman scaling \citep{guttman_principal_1950}. In the latter, the items in a scale form a strict hierarchy. If a subject has an item on the scale, then all lower items also apply. A subject's score on the scale is thus the highest item that is true for the subject. 

Intuitively, DLS has these properties. If a language has a good virtual assistant (like Siri), then we can infer that it also has good machine translation---but having good machine translation does not imply having a good virtual assistant.  Similarly, if a language has good machine translation, we can guess that it must also have good spell checking, though we cannot assume that the reverse would hold.  In a Guttman scale, an exception to the hierarchical ordering is considered an error, but in an arena like DLS we can expect there to be exceptions.  Mokken scaling is a method for placing the items of a supposed hierarchical scale into their optimal order, while providing metrics that allow one to evaluate how well the hierarchical model fits. 

\subsection{Categories of Digital Language Support}

The method uses the following seven categories of DLS. They are listed below from easiest (most commonly supported) to hardest (least commonly supported) as determined by the results of our analysis:\footnote{This aspect of the analysis is explained in subsection \ref{difficulty} and illustrated in Figure \ref{fig:difficulty}.}

\begin{itemize}[noitemsep]
 \item Content --- A service offering content in many languages (like Wikipedia, news sites, or Bible sites)\footnote{Having digital content in a language could also be viewed as an evidence of digital language use. We treat the fact that a service offers content in a language as a Boolean indicator of support for the language. To measure  digital language use, we would quantify the amount of digital content in each language.}
 \item Encoding --- A system component for representing languages (like keyboards and fonts)
 \item Surface  --- A tool with surface-level processing (like  spell checking or stemming)
 \item Localized --- A tool with a localized user interface (like operating system, browser, or messaging) 
 \item Meaning --- A tool with meaning-level processing (like machine translation)
 \item Speech --- A tool for speech processing (like speech-to-text or text-to-speech)
 \item Assistant --- An intelligent virtual assistant (like Siri or Alexa)
\end{itemize}

For each category, we sought to identify the top ten tools of its kind globally. In order to ensure that we included the major tools in use outside the English-speaking world, we also included the top five tools in each of the ten most populous countries of the world.\footnote{This sampling method allows us to discover widely-used tools that support just one large language, but it admittedly misses tools that have been custom-built for a single smaller language.}  The reference authority for these rankings was the \textit{similarweb} service.\footnote{\url{https://similarweb.com}} Then we added any tools found from other sources that  supported more than 10\% of the median number of languages supported by the top tools in the category.
In order for a tool to be used in our analysis, we required there to be a URL from which the names or ISO 639 codes of supported languages could be scraped. 

The full sample consists of URLs for 143 digital tools across the seven categories of DLS.\footnote{A complete list of the 143 digital tools is provided at  \url{https://github.com/sil-ai/dls-results}.} The number of tools in each category is shown in Table~\ref{tbl:subscales} as the  maximum number in the range for level~4.

\subsection{Harvesting the feature data}

The method works by scraping each URL in the sample to discover what languages each tool supports.  The harvested language names are mapped to their corresponding ISO 639-3 code\footnote{\url{https://iso639-3.sil.org/code_tables}} by means of a manually maintained table of name-to-code mappings. After the mapping of the harvested language names, the resulting feature data is a logical matrix  with rows for 7,829 ISO 639-3 codes, columns for the 143 digital tools, and a Boolean value at the intersection indicating whether the given language is supported by the given tool.   

\subsection{Scoring the DLS categories as subscales}

When a language is not supported by any tools in a given DLS category it is scored as 0; otherwise, the  number of tools supporting that language is converted to a level score on a four-level subscale. The correspondence between the number of tools supporting the language and the level on the subscale is shown in Table \ref{tbl:subscales}. The score corresponds to the quartile in the distribution of the number of tools supporting each language; only the languages that are supported by at least one tool in the category are included in that distribution.\footnote{The quartile boundaries are extended upward to accommodate ties; thus in every case, Level 1 contains more than 25\% of the languages with that kind of support.}

\begin{table}[htb]
\centering
\begin{tabular}{l | c c c c c}
\hline
\textbf{Category} & \multicolumn{4}{c}{\textbf{Levels}} \\
\textbf{} & \textbf{1} & \textbf{2} & \textbf{3} & \textbf{4} \\
\hline
Assistant & 1 & 2 & 3--4 & 5--11 \\
Speech & 1 & 2--3 & 4--8 & 9--23 \\
Meaning & 1 & 2 & 3--6 & 7--14 \\
Localized & 1 & 2 & 3--12 & 13--47 \\
Surface & 1 & 2 & 3 & 4--15 \\
Encoding & 1 & 2 & 3 & 4--10 \\
Content & 1 & 2 & 3 & 4--23 \\
\hline
\end{tabular}
\caption{\label{tbl:subscales}
Number of tools supporting a language in each level of the subscales for the DLS categories
}
\end{table}

\subsection{From category levels to scale items}

\begin{figure}[htb]
  \centering
  \includegraphics[width=\columnwidth]{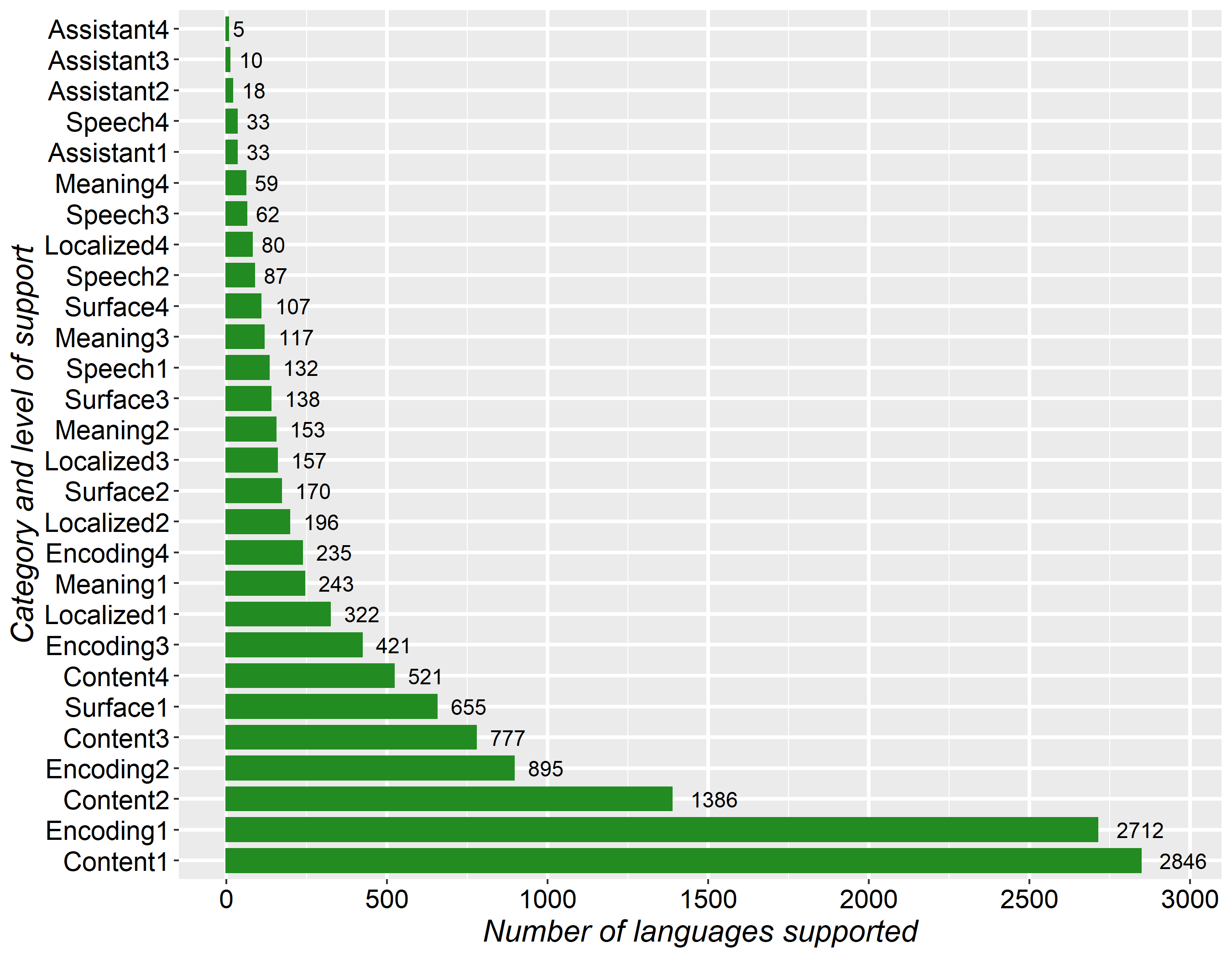}
  \caption{Number of languages supported at each category and level of digital language support }
  \label{fig:scale}
\end{figure}

In constructing the Mokken scale, the levels of the categories become items in the scale. These items are named Content1, Content2, and so on. Within each subscale, the items form a strict hierarchy, in which being scored at a higher level on the subscale implies also having at least as much support as the lower levels of the same subscale.  Thus the count of languages for item Content3 also includes the languages for Content4, and so on going down. 
The bar graph in Figure \ref{fig:scale} shows the items listed from top to bottom in ascending order of the number of languages with at least that level of support in the named category. 

\section{Results}

\subsection{Evaluating fit of the model}

Mokken scale analysis allows us to evaluate the degree to which the scale depicted in Figure \ref{fig:scale} forms a hierarchical scale.  This is done using Loevinger's \citeyearpar{loevinger_technic_1948} coefficient of homogeneity, \textit{H}.\footnote{We have
performed these calculations using the  “mokken” package \citep{ark_mokken_2007,ark_new_2012} in R \citep{team_r_2022}.} 
\textit{H} compares the actual Guttman errors to the expected number of errors if the items were not related in a scale.  A value of 1.0 indicates no errors; any value above 0.5 is indicative of a strong scale \citep{sijtsma_introduction_2002}.

\begin{table}[htb]
\centering
\begin{tabular}{lc}
\hline
\textbf{Item} & \textbf{\textit{H}}\\
\hline
Assistant &	0.987 \\
Speech &	0.942 \\
Meaning &	0.920 \\
Localized &	0.924 \\
Surface &	0.885 \\
Encoding &	0.707 \\
Content &	0.685 \\
\hline
Full scale &	0.825 \\
\hline
\end{tabular}
\caption{Coefficient of homogeneity, \textit{H}, for DLS scale}
\label{tab:HCoefs}
\end{table}

The results in Table \ref{tab:HCoefs} show that the proposed DLS scale is a very strong scale, especially among the categories of support that are hardest to achieve. Thus the total score on all 7 categories (i.e., 0 to 28) serves to quantify the DLS for a given language.

\subsection{Relative difficulty of DLS items}\label{difficulty}

\begin{figure}[htb]
  \centering
  \includegraphics[width=\columnwidth]{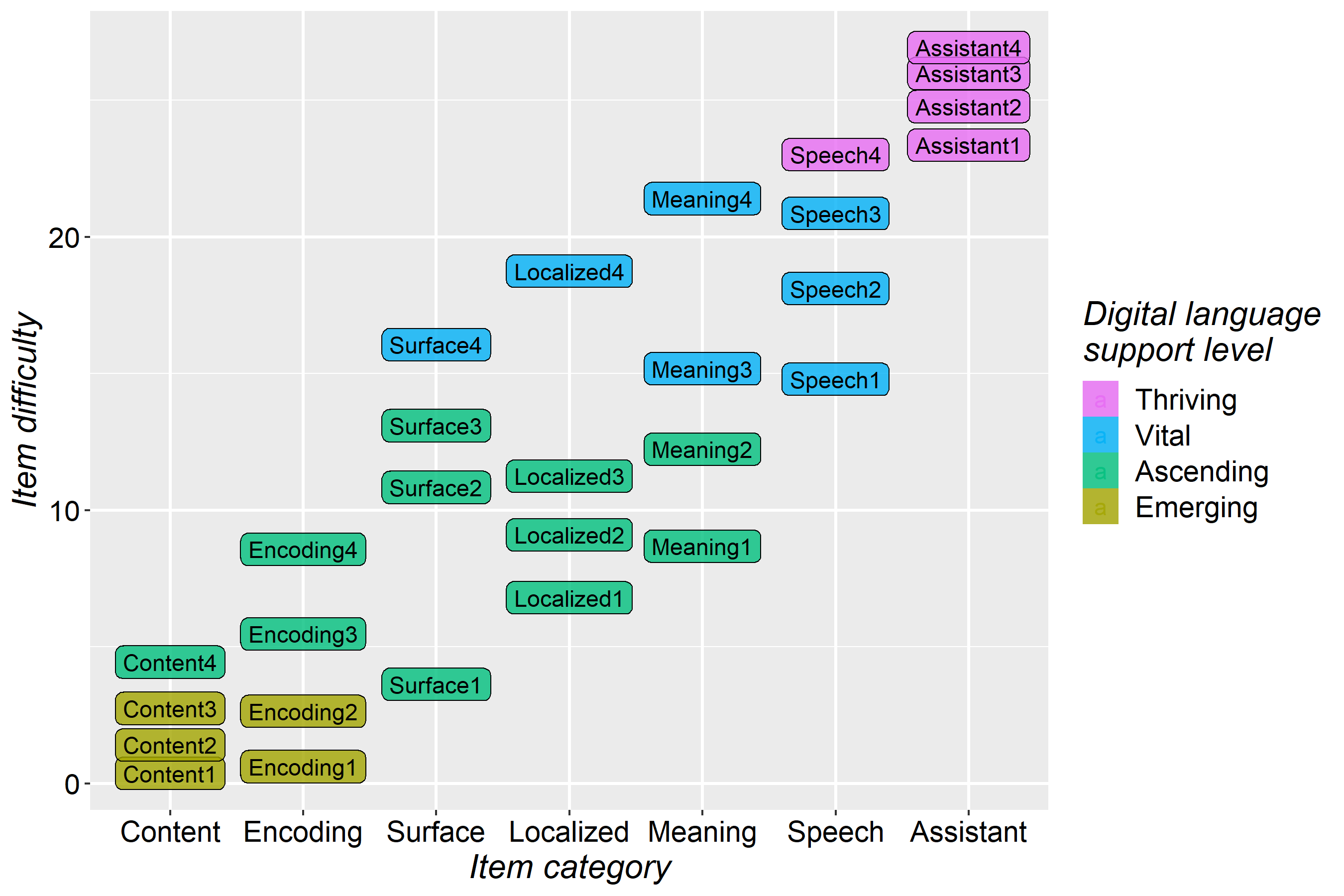}
  \caption{Difficulty of the DLS categories and levels}
  \label{fig:difficulty}
\end{figure}

Mokken analysis is based on Item Response Theory (IRT)---a methodology developed for educational and psychological testing \citep{lord_applications_1980}. 
In IRT, logistic regression is used to derive an Item Response Function (IRF) for each test item; it returns the probability that a subject would produce a positive (or correct) 
response on that item, given their total score on the rest of the test items. The difficulty of an item is defined as the  score (on the rest of the test) at which the subject has a 50\% chance of giving a positive response for the item. Figure \ref{fig:difficulty} plots the difficulty for each of the scale items listed in Figure \ref{fig:scale}.  For instance, a language has a 50\% chance of getting its first spell-checker (Surface1) if it has 3.6 other DLS items, but the first virtual assistant (Assistant1) cannot be expected until it has 23.4 other DLS items.

\subsection{DLS as a growth curve}

\begin{figure}[htb]
  \centering
  \includegraphics[width=\columnwidth]{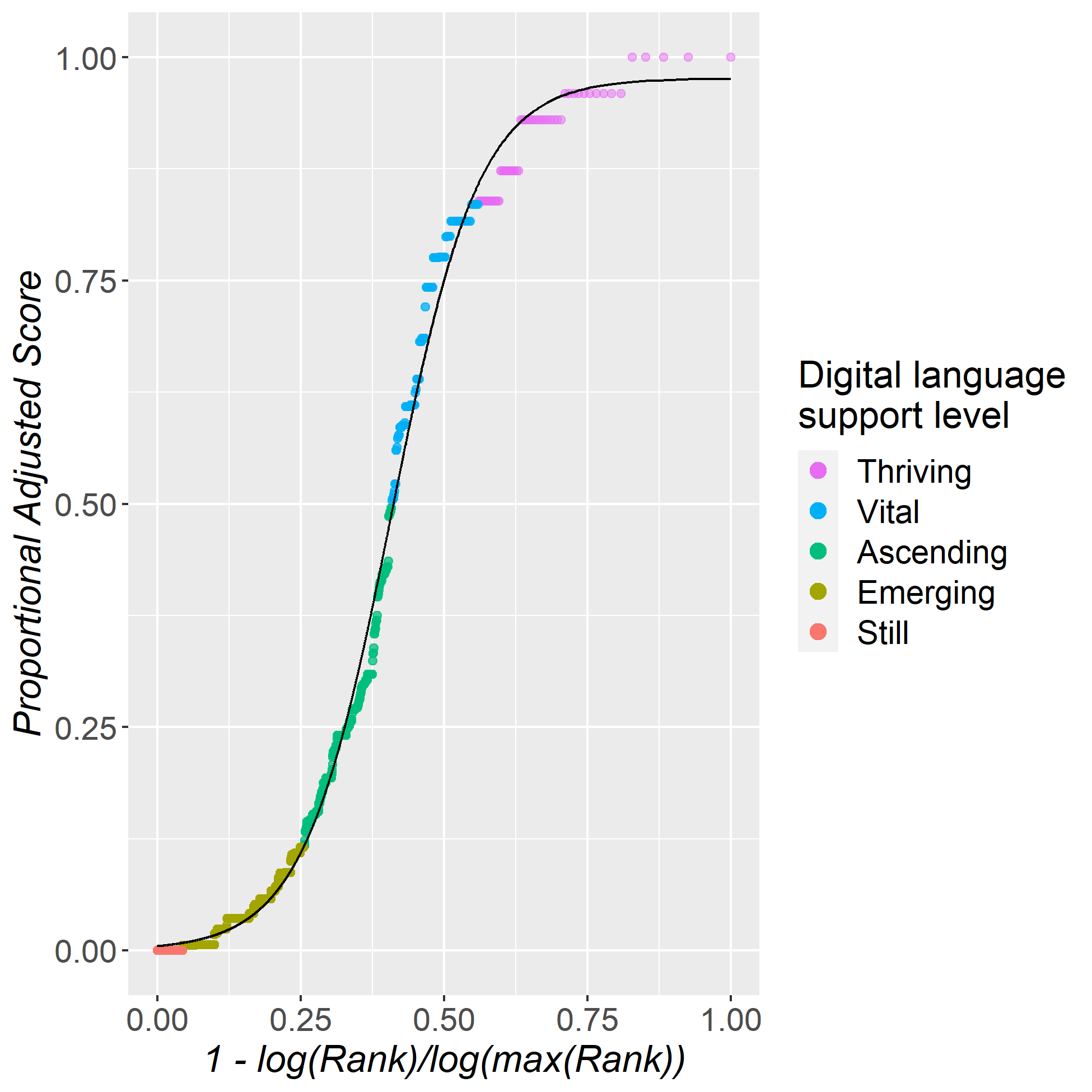}
  \caption{The growth of DLS as a logistic function. }
  \label{fig:sCurve}
\end{figure}

Figure \ref{fig:sCurve} plots the DLS score for 7,829 ISO 639 languages. The vertical axis is the measure of DLS as a proportion:  the DLS score achieved divided by the maximum possible score.\footnote{The DLS scores are also adjusted by scoring each item as the probability returned by its IRF.  In educational testing, scoring each positive response as a probability is a way of controlling for random guessing on
questions that are too hard for the subject. In the application to DLS it can control for "random" developments that do not have the underpinnings of the expected lower categories of support, such as when there is a one-time philanthropic gesture by a large company or the potentially unsustainable efforts of a solitary developer.}
The horizontal axis is the rank of the language by DLS score, 
but converted to a log scale and flipped so that lowest DLS is on the left and highest is on the right.
 
The pattern that emerges is an S-curve as is typical in studies of growth in innovation. We follow the geometry of the fitted curve to assign each language to one of the five summary levels: 

\begin{itemize}[noitemsep]
 \item Still --- a score of 0
 \item Emerging --- at the bottom where the slope is  more horizontal than vertical
 \item Ascending  --- below the midpoint where the slope is  more vertical than horizontal
 \item Vital --- above the midpoint where the slope is  more vertical than horizontal  
 \item Thriving --- at the top where the slope is  more horizontal than vertical
\end{itemize}

\noindent By comparing Figures \ref{fig:difficulty} and \ref{fig:sCurve} one sees what components of DLS correspond to the summary levels.

Table \ref{tab:Levels} reports the number of languages at each summary level along with the names of example languages, the first being from the upper end of the range and the second from the lower.\footnote{A sampling of the detailed results produced by the system is provided at  \url{https://github.com/sil-ai/dls-results}.} 

\begin{table}[htb]
\centering
\begin{tabular}{l c l}
\hline
\textbf{Level} & \textbf{Languages} & \textbf{Examples}\\
\hline
Thriving & 33 & English, Hungarian \\
Vital &	95 &  Nepali, Tongan \\
Ascending &	401 &  Greenlandic, Hunsrik \\
Emerging & 3304 &  Dogri, Michif\\
Still &	3996 &  Aimaq, Yurok\\
\hline
\end{tabular}
\caption{Number of languages per DLS level}
\label{tab:Levels}
\end{table}

\section{Conclusion}

%we present x which generally showed y. in the future we'd like to z.
We have presented a method that produces an explainable model for quantifying DLS. We are currently working with Ethnologue to add reporting on DLS in its description of languages, beginning with the next edition. Regularly updating the assessments should serve to document the digital trajectory of every known language.

\section*{Acknowledgements}

This research has been funded by the Ethnologue program of SIL International for the purpose of developing a way to add digital language vitality to its reporting on the state of the world's languages. At the outset of the project,
SIL International licensed the software developed by \citet{kornai_digital_2013} from the Hungarian Academy of Sciences. We are deeply indebted to Andras Kornai and his PhD student at the time, Katalin Pajkossy, for their help and encouragement as we first replicated his results before developing the method described in this paper. Others who made significant contributions include Steve Moitozo, Erica Oldaker (n\'ee Swindle), Steve Woolston, and Daniel Whitenack.

\bibliography{DigitalLanguageVitality.bib}
\bibliographystyle{acl_natbib}

\end{document}